\documentclass{bmvc2k}

\usepackage{amsfonts}
\usepackage{amsmath}
\usepackage{enumitem}
\usepackage{array}
\usepackage[table]{xcolor}

\makeatletter
\RequirePackage[colorlinks,urlcolor=blue,citecolor=red,bookmarks=false]{hyperref}
\bmv@pdfcompresslevel 
\makeatother

\title{Learning to Generate and Reconstruct 3D Meshes with only 2D Supervision}

\addauthor{Paul Henderson}{p.m.henderson@ed.ac.uk}{1}
\addauthor{Vittorio Ferrari}{vittoferrari@google.com}{2}

\addinstitution{
School of Informatics\\University of Edinburgh\\Edinburgh, Scotland
}
\addinstitution{
Google Research\\Z\"{u}rich, Switzerland
}

\runninghead{Henderson \& Ferrari}{Mesh Generation \& Reconstruction}

\newcommand{\sect}[1]{Sec.~\ref{sec:#1}}
\newcommand{\fig}[1]{Fig.~\ref{fig:#1}}
\newcommand{\tab}[1]{Table~\ref{tab:#1}}
\newcommand{\eqn}[1]{(\ref{eq:#1})}

\newcommand{\given}{\,|\,}
\newcommand{\Given}{\,\Big\vert\,}
\newcommand{\kldiv}[2]{\mathit{KL}\left[ #1 \,\Big\rvert\Big\rvert\, #2 \right]}

\setlist{itemsep=0pt,topsep=1pt,parsep=1pt,partopsep=2pt,leftmargin=16pt}

\newcommand{\setlengths}{
\setlength{\belowcaptionskip}{6pt}%
\setlength{\abovedisplayskip}{3pt}%
\setlength{\belowdisplayskip}{3pt}%
\setlength{\abovedisplayshortskip}{3pt}%
\setlength{\belowdisplayshortskip}{3pt}%
\setlength{\jot}{2pt}
}

\makeatletter
\renewcommand{\paragraph}{%
  \@startsection{paragraph}{4}%
  {\z@}{0.4ex \@plus 0.0ex \@minus 0.15ex}{-0.4em}%
  {\normalfont\normalsize\bfseries}%
}
\makeatother

\begin{document}

\setlengths
\maketitle

\begin{abstract}
We present a unified framework tackling two problems: class-specific 3D reconstruction from a single image, and generation of new 3D shape samples.
%
These tasks have received considerable attention recently; however, existing approaches rely on 3D supervision, annotation of 2D images with keypoints or poses, and/or training with multiple views of each object instance.
Our framework is very general:  it can be trained in similar settings to these existing approaches, while also supporting weaker supervision scenarios. Importantly, it can be trained purely from 2D images, without ground-truth pose annotations, and with a single view per instance.
%
We employ meshes as an output representation, instead of voxels used in most prior work.
This allows us to exploit shading information during training, which previous 2D-supervised methods cannot.
Thus, our method can learn to generate and reconstruct concave object classes.
%
%
%
%
We evaluate our approach on synthetic data in various settings, showing that
(i) it learns to disentangle shape from pose;
(ii) using shading in the loss improves performance;
(iii) our model is comparable or superior to state-of-the-art voxel-based approaches on quantitative metrics, while producing results that are visually more pleasing;
(iv) it still performs well when given supervision weaker than 
in
prior works. 
\end{abstract}


\section{Introduction}
\label{sec:intro}

\begin{figure}
  \includegraphics[width=\linewidth]{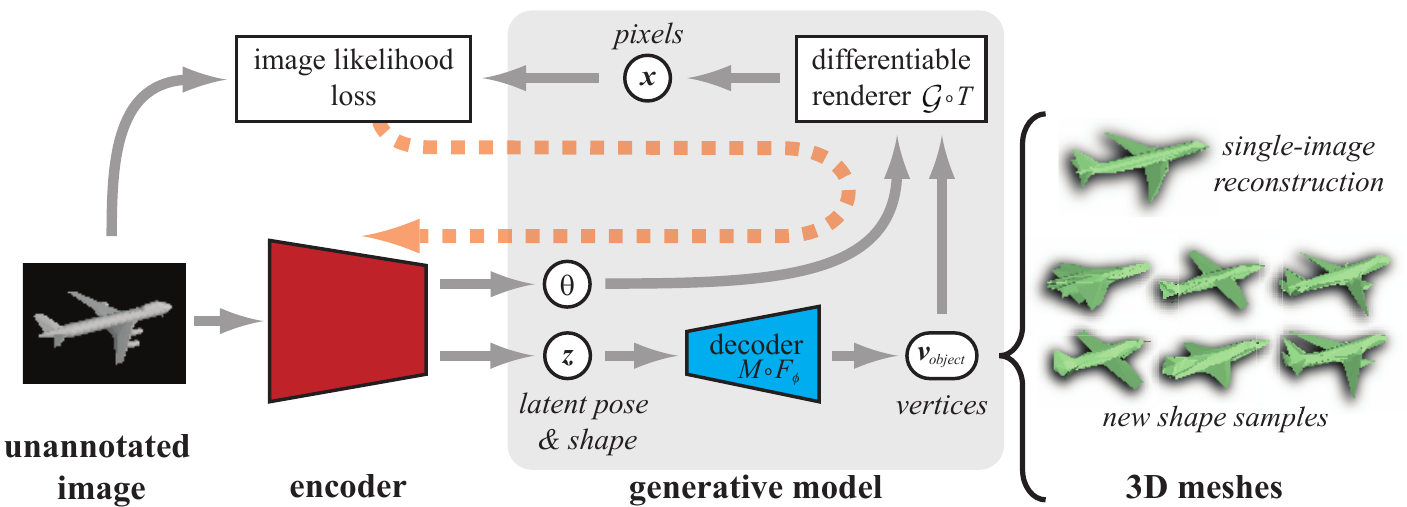}
  \vspace{-20pt}
  \caption{
Given only unannotated 2D images as training data, our model learns (1) to reconstruct and predict the pose of 3D meshes from a single test image, and (2) to generate new 3D mesh samples.
It is trained end-to-end (orange dashed arrow) to reconstruct input images, via a differentiable renderer that produces lit, shaded RGB images, allowing us to exploit shading cues in the loss.
}
  \label{fig:flow}
\end{figure}

Reconstructing 3D objects from 2D images is a long-standing research area in computer vision.
While traditional methods rely on multiple images of the same object instance~\cite{seitz2006comparison,furukawa15cgft,broadhurst01iccv,laurentini94pami,debonet99iccv,gargallo07accv,liu10cvpr}, there has recently been a surge of interest in learning-based methods that can infer 3D structure from a single image, assuming that it shows an object of a class seen during training~\cite{fan17cvpr,zhu17iccv-rethinking,gwak173dv,choy16eccv,yan16nips,girdhar16eccv,tulsiani17cvpr,wiles17bmvc}.
A related problem to reconstruction is that of generating new 3D shapes from a given object class \textit{a priori}, i.e. without conditioning on an image.
Again, there have recently been several works that apply deep learning techniques to this task~\cite{wu16nips,gadelha173dv,sinha17cvpr,zou17iccv,li17tog}.

Most learning-based methods for reconstruction and generation rely on strong supervision.
For generation, \cite{wu16nips,rezende16nips,sinha17cvpr,zou17iccv,li17tog} use large collections of manually constructed 3D shapes~\cite{shapenet15arxiv,wu15cvpr-shapenets}.
For reconstruction, \cite{girdhar16eccv,choy16eccv,fan17cvpr,zhu17iccv-rethinking} require training images paired with aligned 3D meshes; \cite{gwak173dv} relaxes this slightly by not requiring the images and meshes to be paired.
While other methods do not rely on 3D ground-truth, they still require annotations on the 2D training images such as keypoints~\cite{kar15cvpr,vicente14cvpr} and object poses \cite{tulsiani17cvpr,yan16nips,wiles17bmvc}. Furthermore, some of them also require multiple views for each object instance \cite{yan16nips,wiles17bmvc,rezende16nips}. 
In this paper, we consider the more challenging setting where we only have access to unannotated 2D images for training, without ground-truth pose, keypoints, or 3D shape, and with a single view per object instance; this setting is considered in just one previous work~\cite{gadelha173dv}.


It is well known that \textit{shading} provides an important cue for 3D understanding~\cite{horn75pcv}.
It allows determination of surface orientations, if the lighting and material characteristics are known; this has been explored in numerous works on shape-from-shading over the years~\cite{horn75pcv,zhang99pami,barron15pami}.
Unlike learning-based approaches, these methods can only reconstruct non-occluded parts of an object, and achieving good results requires strong priors~\cite{barron15pami}.
Conversely, existing learning-based generation and reconstruction methods can reason over occluded or visually-ambiguous areas, but do not leverage shading information in their loss.
Furthermore, the vast majority use voxel grids as an output representation (except \cite{zou17iccv,sinha17cvpr}); while easy to work with, these cannot model surfaces that are not axis-aligned, limiting the usefulness of shading cues.
To exploit shading information in a learning-based approach, we therefore need to move beyond voxels; a natural choice of representation is then 3D \textit{meshes}.
Meshes are ubiquitous in computer graphics, and have desirable properties for our task: they can represent surfaces of arbitrary orientation and dimensions at fixed cost, and are able to capture fine details.
Thus, they avoid the visually displeasing `blocky' reconstructions that result from voxels.
We also go beyond monochromatic light, considering the case of coloured directional lighting; this provides even stronger shading cues when combined with arbitrarily-oriented mesh surfaces.

In this paper, we present a unified framework for both reconstruction and generation of 3D shapes, that is trained with only 2D supervision, and models 3D meshes rather than voxels (\fig{flow}).
Our framework is very general, and can be trained in similar settings to existing models~\cite{tulsiani17cvpr,yan16nips,wiles17bmvc}, while also supporting weaker supervision scenarios.
It allows:
\begin{itemize}
\item
use of different \textbf{mesh parameterisations}, which lets us incorporate useful modeling priors such as smoothness or composition from primitives
\item
exploitation of \textbf{shading cues} due to monochromatic or coloured directional lighting, letting us discover concave structures that silhouette-based methods~\cite{gadelha173dv,tulsiani17cvpr,yan16nips} cannot
\item
training with \textbf{varying degrees of supervision}: single or multiple views per instance, with or without ground-truth pose annotations
\end{itemize}

To achieve this, we design a probabilistic generative model that captures the full image formation process, whereby the shape and pose of a 3D mesh are first sampled independently, then a 2D rendering is produced from these (\sect{generative}).
We use stochastic gradient variational Bayes~\cite{kingma14iclr,rezende14icml} for training (\sect{training}).
This involves learning an \textit{inference network} that can predict 3D shape and pose from a single image, with the shape placed in a canonical frame of reference, i.e. disentangled from the pose.
Together, the model plus its inference network resemble a variational autoencoder~\cite{kingma14iclr} on pixels.
It represents 3D shapes in a compact latent embedding space, and has extra layers in the decoder corresponding to the mesh representation and renderer.
%
As we do not provide 3D supervision, the encoder and decoder must bootstrap and guide one another during training. The decoder learns the manifold of shapes, while at the same time the encoder learns to map images onto this.
This learning process is driven purely by the objective of reconstructing the training images.
While this is an ambiguous task and the model cannot guarantee to reconstruct the true shape of an object from a single image, its generative capability means that it always produces a plausible instance of the relevant class; the encoder ensures that this is consistent with the observed image.
%
This works because
the generative model must learn to produce shapes that reproject well over {\em all} training images, starting from low-dimensional latent representations.
This creates an inductive bias towards regularity, which avoids degenerate solutions with unrealistic shapes that could, in isolation, explain each individual training image.

We display samples from our model in \sect{gen-results}, showing that (i) the use of meshes yields more natural samples than those from voxel-based methods, and (ii) our samples are diverse and realistic.
In \sect{recon-results}, we quantitatively evaluate the performance of our method on single-view reconstruction and pose estimation, in the various settings described above.
We show that
(i) it learns to predict pose, and disentangle it from shape;
(ii) exploiting information from shading improves the results;
(iii) it achieves comparable or better performance than prior works with equivalent supervision; and
(iv) it still performs well when given weaker supervision than supported by prior works.

\begin{figure}
  \includegraphics[width=\linewidth]{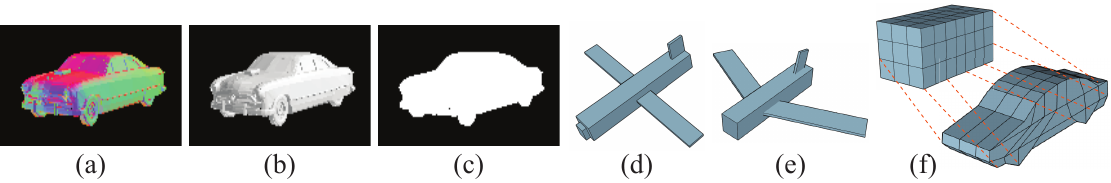}
  \vspace{-22pt}
  \caption{
\textbf{Lighting}: Coloured directional lighting (a) provides strong cues for surface orientation; white light (b) provides less information; silhouettes (c) provide none at all. Our model is able to exploit the shading information from coloured or white lighting.
\textbf{Mesh parameterisations}: \textbf{ortho-block} \& \textbf{full-block} (assembly from cuboidal primitives, of fixed or varying orientation) are suited to objects consisting of compact parts (d-e); \textbf{subdivision} (per-vertex deformation of a subdivided cube) is suited to complex continuous surfaces (f).
}
  \label{fig:light-and-paramns}
\end{figure}

\section{Generative Model}
\label{sec:generative}

Our goal is to build a probabilistic generative model of 3D meshes for a given object class.
For this to be trainable with 2D supervision, we cast the entire image-formation process as a directed model (\fig{flow}).
We assume that the content of an image can be explained by two independent latent components---the shape of the mesh, and its pose relative to the camera.
These are modelled by two low-dimensional random variables, $\mathbf{z}$ and $\theta$ respectively. The joint distribution over these and the resulting pixels $\mathbf{x}$ factorises as
$
  P(\mathbf{x}, \mathbf{z}, \theta)
   =
  P(\theta) P(\mathbf{z}) P(\mathbf{x} \given \mathbf{z}, \theta)
$.

Following \cite{gadelha173dv,yan16nips,tulsiani17cvpr,wiles17bmvc}, we assume that the pose $\theta$ is parameterised by just the azimuth angle, with $\theta \sim \text{Uniform}(-\pi, \pi)$.
The camera is then placed at fixed distance and elevation relative to the object.
Following recent works on deep latent variable models~\cite{kingma14iclr,goodfellow14nips}, we assume that $\mathbf{z}$ is drawn from a standard isotropic Gaussian, and then transformed by a deterministic \textit{decoder network},
 $F_{\phi}$, parameterised by weights $\phi$ which are to be learnt.
This produces the \textit{mesh parameters} $\Pi = F_{\phi}(\mathbf{z})$.
Intuitively, the decoder network $F_{\phi}$ transforms and entangles the dimensions of $\mathbf{z}$ such that all values in the latent space map to plausible values for $\Pi$, even if these lie on a highly nonlinear manifold.
Note that our approach contrasts with previous models that directly output pixels~\cite{kingma14iclr,goodfellow14nips} or voxels~\cite{wu16nips,gadelha173dv} from a decoder network.

We use $\Pi$ as inputs to a fixed mesh parameterisation function $M(\Pi)$, which yields vertices $\mathbf{v}_{\text{object}}$ of triangles defining the shape of the object in 3D space, in a canonical pose (different options for $M$ are described below).
The vertices are transformed into camera space according to the pose $\theta$, by a fixed function $T$: $\mathbf{v}_{\text{camera}} = T(\mathbf{v}_{\text{object}},\, \theta)$.
They are then rendered into an RGB image $I_0 = \mathcal{G}(\mathbf{v}_{\text{camera}})$ by a rasteriser $\mathcal{G}$ with Gouraud shading~\cite{gouraud71tc} and Lambertian directional lighting~\cite{lambert60bk}.
We are free to choose the lighting parameters: our experiments include tri-directional coloured lighting, and white directional lighting with an ambient component.

The final observed pixel values $\mathbf{x}$ are modelled as independent Gaussian random variables, with means equal to the values in an $L$-level Gaussian pyramid~\cite{burt83tc}, whose base level equals $I_0$, and whose $L$\textsuperscript{th} level has smallest dimension equal to $1$:
\begin{align}
    P_{\phi}(\mathbf{x} \given \mathbf{z}, \theta) &= \prod_l P_{\phi}(\mathbf{x}_l \given \mathbf{z}, \theta) 
  &
    \mathbf{x}_l &\sim \text{Normal}\left( I_l, \tfrac{\epsilon}{2^l} \right)
  \\
    I_0 &= \mathcal{G}(T(M(F_{\phi}(\mathbf{z})),\, \theta))
  &
    I_{l+1} &= I_l * k_G
\end{align}
where $k_G$ is a small Gaussian kernel, $\epsilon$ is the noise magnitude at the base scale, and $*$ denotes convolution with stride two.
We use a multi-scale pyramid instead of just the raw pixel values to ensure that, during training, there will be gradient forces over long distances in the image, thus avoiding bad local minima where the reconstruction is far from the input.

\paragraph{Mesh parameterisations.}
After the decoder network has transformed the latent embedding $\mathbf{z}$ into the mesh parameters $\Pi$, these are converted to actual 3D vertices using a simple, non-learnt mesh-parameterisation function $M$.
One possible choice for $M$ is the identity function, in which case the decoder network directly outputs vertex locations.
However, initial experiments showed that this does not work well: it produces very irregular meshes with large numbers of intersecting triangles.
Conversely, using a more sophisticated form for $M$ enforces regularity of the mesh.
We use three different parameterisations in our experiments.

In our first parameterisation, $\Pi$ specifies the locations and scales of a fixed number of axis-aligned cuboidal \textit{primitives} (\fig{light-and-paramns}d), from which the mesh is assembled~\cite{zou17iccv,tulsiani17cvpr-blocks}.
Changing $\Pi$ can produce configurations with different topologies, depending which blocks touch or overlap, but all surfaces will necessarily be axis-aligned.
In our experiments we call this \textbf{ortho-block}.

Our second parameterisation is strictly more powerful than the first: we still assemble the mesh from cuboidal primitives, but now parameterise each with a 3D rotation, in addition to its location and scale.
In our experiments we call this \textbf{full-block} (\fig{light-and-paramns}e).

The above parameterisations are naturally suited to objects composed of compact parts, but cannot represent complex continuous surfaces.
For these, we define a third parameterisation, \textbf{subdivision} (\fig{light-and-paramns}f).
This parameterisation is based on a single unit cube, centred at the origin; the edges and faces of the cube are subdivided several times along each axis.
Then, $\Pi$ specifies a list of displacements, one per vertex, which deform the subdivided cube into the required shape.

\section{Variational Training}
\label{sec:training}

We wish to learn the parameters of our model from a training set of 2D images of objects of a single class.
More precisely, we assume access to a set of images $\{\mathbf{x}^{(i)}\}$, each showing an unknown object instance at unknown pose.
Note that we do \textit{not} require that there are multiple views of each object (in contrast with \cite{yan16nips}), nor that the object poses are given as supervision (in contrast with \cite{yan16nips,tulsiani17cvpr,wiles17bmvc}).

We seek to maximise the marginal log-likelihood of the training set, which is given by $\sum_i \log P_{\phi}(\mathbf{x}^{(i)})$, with respect to $\phi$.
For each image, we have
\begin{equation}
\label{eq:loglik}
\log P_{\phi}(\mathbf{x}^{(i)}) = \log \int_{\mathbf{z},\theta} P_{\phi}(\mathbf{x}^{(i)} \given \mathbf{z}, \theta) P(\mathbf{z}) P(\theta) d\mathbf{z} \, d\theta.
\end{equation}
Unfortunately this is intractable, due to the integral over the latent space $\mathbf{z}, \theta$.
Hence, we use amortised variational inference, in the form of stochastic gradient variational Bayes \cite{kingma14iclr,rezende14icml}.
This introduces an approximate posterior $Q_{\omega}(\mathbf{z}, \theta \given \mathbf{x})$, parameterised by some $\omega$ that we learn jointly with the model parameters $\phi$.
Intuitively, $Q$ maps an image $\mathbf{x}$ to a distribution over likely values of the latent variables $\mathbf{z}$ and $\theta$.
Instead of the log-likelihood \eqn{loglik}, we then maximise the \textit{evidence lower bound} (ELBO):
\begin{equation}
\label{eq:elbo}
  \mathop{\mathbb{E}}_{\mathbf{z}, \theta \sim Q_{\omega}(\mathbf{z}, \theta \given \mathbf{x}^{(i)})}\left[ 
    \log P_{\phi}( \mathbf{x}^{(i)} \given \mathbf{z}, \theta )
  \right] - \kldiv{
    Q_{\omega}(\mathbf{z}, \theta \given \mathbf{x}^{(i)})
  }{
    P(\mathbf{z}) P(\theta)
  }
  \le
  \log P_{\phi}(\mathbf{x}^{(i)}).
\end{equation}
This lower-bound on the log-likelihood can be evaluated efficiently, as the necessary expectation is now with respect to $Q$, for which we are free to choose a tractable form.
The expectation can then be approximated using a single sample.

We let $Q$ be a mean-field approximation, factorised as $Q_{\omega}(\mathbf{z}, \theta \given \mathbf{x}) = Q_{\omega}(\mathbf{z} \given \mathbf{x}) Q_{\omega}(\theta \given \mathbf{x})$.
$Q_{\omega}(\mathbf{z} \given \mathbf{x})$ is a multivariate Gaussian with diagonal covariance.
The mean and variance of each latent dimension are given by an \textit{encoder network}, $\mathrm{enc}_{\omega}(\mathbf{x})$, which takes the image $\mathbf{x}$ as input.
For this encoder network we use a CNN with architecture similar to \cite{wiles17bmvc}.
When training with multiple views per instance, we apply the encoder network to each image separately, then calculate the final shape embedding $\mathbf{z}$ by max-pooling each dimension over all views.

For the pose $\theta$, we could similarly use a Gaussian posterior.
However, many objects are roughly symmetric with respect to rotation, and so the true posterior is typically multi-modal.
We capture this multi-modality by decomposing the rotation into coarse and fine parts~\cite{mousavian17cvpr}: an integer random variable $\theta_{\text{coarse}}$ that chooses from $R$ rotation bins, and a small Gaussian offset $\theta_{\text{fine}}$ relative to this.
We apply this transformation in both the generative $P(\theta)$ and variational $Q_{\omega}(\theta)$, giving
\begin{equation}
  \theta = -\pi + \theta_{\text{coarse}} \frac{2\pi}{R} + \theta_{\text{fine}}
\end{equation}
\begin{align}
  \label{eq:theta-prior}
  P(\theta_{\text{coarse}} = r) &= 1/R, &\;\;\; P(\theta_{\text{fine}}) &= \text{Normal}(\theta_{\text{fine}} \given 0, \pi / R )
\\
  Q_{\omega}\left( \theta_{\text{coarse}} = r \Given \mathbf{x}^{(i)} \right) &= \rho_r \left( \mathbf{x}^{(i)} \right), &\;\;\; Q_{\omega}(\theta_{\text{fine}}) &= \text{Normal}\left( \theta_{\text{fine}} \Given \xi(\mathbf{x}^{(i)}), \zeta(\mathbf{x}^{(i)}) \right)
\end{align}
where the variational parameters $\rho_r, \xi, \zeta$ for image $\mathbf{x}^{(i)}$ are again estimated by the encoder network $\mathrm{enc}_{\omega}(\mathbf{x}^{(i)})$.
Provided $R$ is sufficiently small, we can integrate directly with respect to $\theta_{\text{coarse}}$ when evaluating \eqn{elbo}, i.e. sum over all possible rotations. We found in initial experiments that this significantly improves performance.

\begin{figure}
  \centering
  \includegraphics[width=0.95\linewidth]{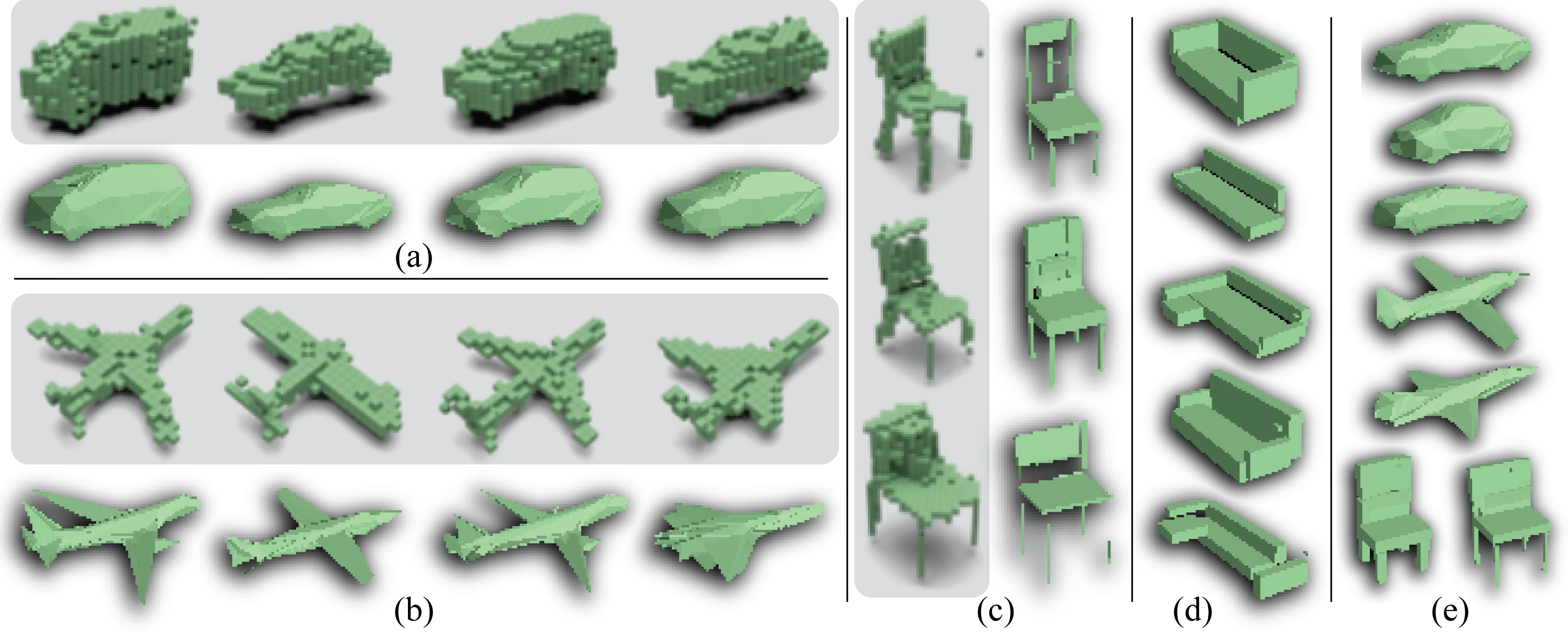}
  \vspace{-10pt}
  \caption{\textbf{(a-c)} Samples from \cite{gadelha173dv} (grey background), shown next to stylistically-similar samples from our model (white background). Both are trained with a single view per instance, and without ground-truth pose. However, our model outputs meshes, and uses shading in the loss. \textbf{(d)} For \textit{sofa}, we only show samples from our model, as \cite{gadelha173dv} cannot handle sofas due to the concavities. \textbf{(e)} Additional samples from our model, showing their diversity and quality}
  \label{fig:samples}
\end{figure}

\paragraph{Imposing a uniform pose prior.}
While the above allows our training process to reason over different poses, it is still prone to predicting the same pose $\theta$ for every image;
clearly this does not correspond to the prior on $\theta$ given by \eqn{theta-prior}.
The model is therefore relying on the shape embedding $\mathbf{z}$ to model all variability, rather than disentangling shape and pose.
The ELBO \eqn{elbo} does include a KL-divergence term that should encourage latent variables to match their prior.
However, it does not have a useful effect for $\theta_{\text{coarse}}$: minimising the KL divergence from a uniform distribution for each sample individually corresponds to independently minimising all the probabilities $Q_{\omega}(\theta_{\text{coarse}})$, which does not encourage uniformity of the full distribution.
The effect we desire is to match the aggregated posterior distribution $\left\langle Q_{\omega}(\theta \given \mathbf{x}^{(i)}) \right\rangle_i$ to the prior $P(\theta)$, where $\langle \,\cdot\, \rangle_i$ is the empirical mean over the training set.
As $\theta_{\text{coarse}}$ follows a categorical distribution in both generative and variational models, we can directly minimise the L1 distance between the aggregated posterior and the prior:
\begin{equation}
\sum_r^R \Big\lvert 
  \left\langle
    Q_{\omega}\left(\theta_{\text{coarse}} = r \given \mathbf{x}^{(i)}\right)
  \right\rangle_i
  - P\left(\theta_{\text{coarse}} = r\right) 
\Big\rvert
=
\sum_r^R \Big\lvert
  \left\langle
    \rho_r(\mathbf{x}^{(i)})
  \right\rangle_i
  - \frac{1}{R} \;
\Big\rvert.
\end{equation}
We use this term in place of $\kldiv{ Q(\theta_{\text{coarse}} \given \mathbf{x}^{(i)}) }{ P(\theta_{\text{coarse}}) }$ in our loss, approximating the empirical mean with a single minibatch.

\paragraph{Loss.}
Our final loss function for a minibatch $\mathcal{B}$ is then given by
\begin{multline}
\label{eq:loss}
  \hspace{-10pt}
  \sum_r^R
  \left\{
    - 
    \left\langle
      \rho_r(\mathbf{x}^{(i)})
      \mathop{\mathbb{E}}_{
        \mathbf{z}, \theta_{\text{fine}} \sim Q_{\omega}
      }\left[ 
        \log P_{\phi}\!\left( \mathbf{x}^{(i)} \Given \mathbf{z}, \theta_{\text{coarse}} = r, \theta_{\text{fine}} \right)
      \right]
   \right\rangle_{\!i \in \mathcal{B}}
   \!
   + \alpha \,
    \Big\lvert
      \!
      \left\langle
        \rho_r(\mathbf{x}^{(i)})
      \right\rangle_{\!i \in \mathcal{B}}
      \!
      - \frac{1}{R} \;
    \Big\rvert
  \right\} \\
  + \beta \,
  \left\langle
    \kldiv{
      Q_{\omega}\left( \mathbf{z}, \theta_{\text{fine}} \Given \mathbf{x}^{(i)} \right)
    }{
      P(\mathbf{z}) P(\theta_{\text{fine}})
    }
  \right\rangle_{\!i \in \mathcal{B}}
\end{multline}
where $\beta$ increases the relative weight of the KL term as in \cite{higgins17iclr}, and $\alpha$ controls the strength of the pose prior matching.
We minimise \eqn{loss} with respect to $\phi$ and $\omega$ using ADAM~\cite{kingma15iclr}, applying the reparameterisation trick~\cite{kingma14iclr,rezende14icml} to handle the Gaussian random variables.

\paragraph{Differentiable rendering.}
Note that optimising \eqn{loss} by gradient descent requires differentiating through the mesh-rendering operation $\mathcal{G}$ used to calculate $P_{\phi}(\mathbf{x} \given \mathbf{z}, \theta)$, to find the derivative of the pixels with respect to the vertex locations and colours.
While computing exact derivatives of $\mathcal{G}$ is very expensive, \cite{loper14eccv} describes an efficient approximation.
We employ a similar technique here, and have made our TensorFlow implementation publicly available\footnote{\textit{DIRT: a fast Differentiable Renderer for TensorFlow}, \url{https://github.com/pmh47/dirt}}.

\begin{figure}
  \centering
  \includegraphics[width=\linewidth]{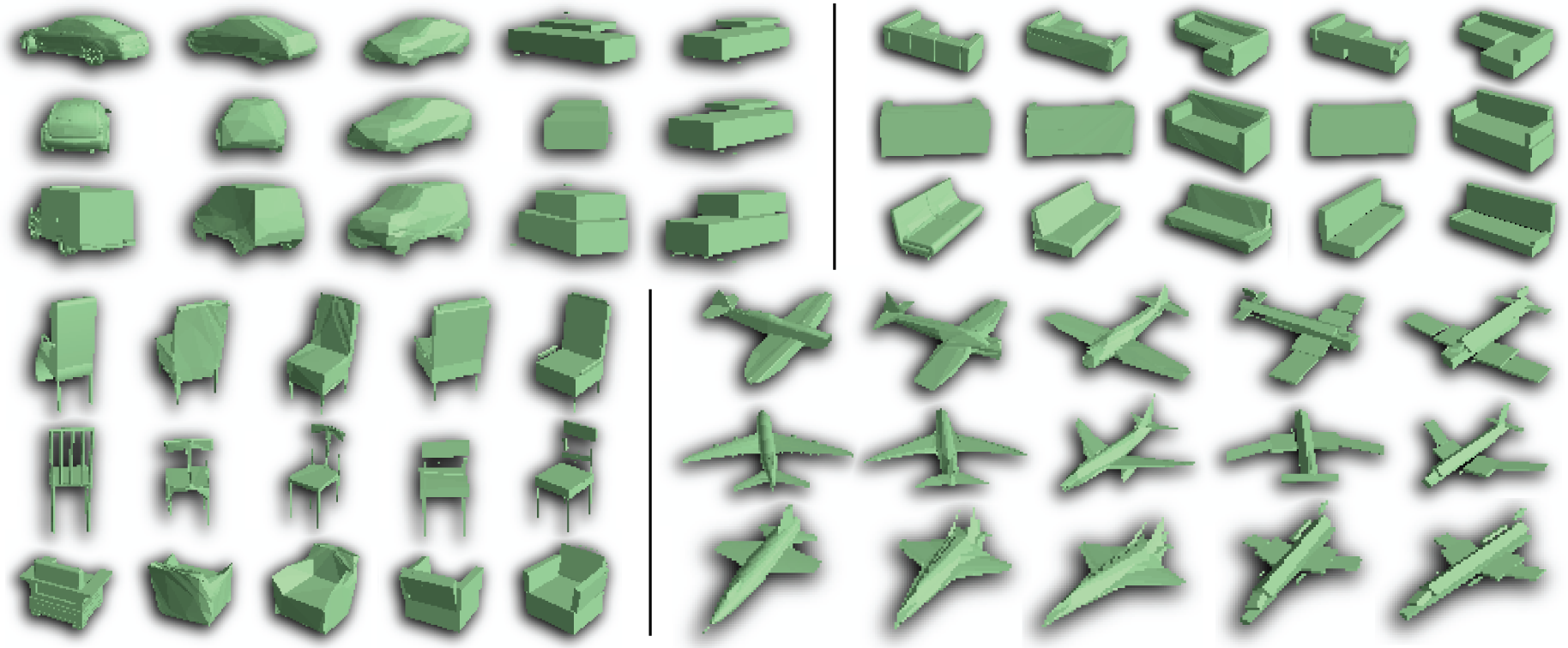}
  \vspace{-20pt}
  \caption{Qualitative examples of reconstructions. Each row of five images shows (i) ShapeNet ground-truth; (ii) our reconstruction with \textbf{subdivision} parameterisation; (iii) reconstruction aligned to canonical pose; (iv) our reconstruction with \textbf{blocks}; (v) aligned reconstruction. Experimental setting: single-view training, colour lighting, shading loss.}
  \label{fig:reconstructions}
\end{figure}

\section{Experiments}

We follow recent works \cite{gadelha173dv,yan16nips,tulsiani17cvpr,fan17cvpr} and evaluate our approach using the ShapeNet dataset~\cite{shapenet15arxiv}.
Using synthetic data has two advantages: it allows
(i) controlled experiments modifying lighting and other parameters; 
(ii) benchmarking the performance of the reconstruction network against ground-truth 3D shapes.
Our experiments focus on the four classes \textit{aeroplane}, \textit{car}, \textit{chair}, and \textit{sofa}.
The first three are used in \cite{gadelha173dv,tulsiani17cvpr,yan16nips}, while the fourth is an example of a highly concave class that is not easily handled by silhouette-based approaches.

To rigorously evaluate the performance of our model, we vary several factors:
\begin{itemize}

\item
\textbf{Mesh parameterisations:} We evaluate the three parameterisations described in \sect{generative}.

\item
\textbf{Lighting:} Unlike previous works~\cite{gadelha173dv,wiles17bmvc,yan16nips,tulsiani17cvpr}, our method is able to exploit shading in the images.
We test in two settings, illumination by
(i) three coloured directional lights (\textbf{colour}), and 
(ii) one white directional light plus a white ambient component (\textbf{white}).

\item
\textbf{Reconstruction loss:}
We typically calculate the reconstruction loss (pixel log-likelihood) over the RGB shaded image (\textbf{shading}), but for comparison with \cite{yan16nips,tulsiani17cvpr,wiles17bmvc} we also experiment with using only the silhouette in the loss (\textbf{silhouette}), disregarding differences in shading between the input and reconstructed pixels.

\item
\textbf{Pose supervision:} Previous works that train for 3D reconstruction with 2D supervision require the ground-truth pose of each training instance~\cite{yan16nips,wiles17bmvc,tulsiani17cvpr}.
Although our method does not need this, we evaluate whether it can benefit from it.

\item
\textbf{Multiple views:} \cite{yan16nips,wiles17bmvc} require that multiple views of each instance are presented together in each training batch, and \cite{tulsiani17cvpr} also focuses on this setting.
Our model does not require this, but for comparison we include results with four views per instance at training time, and either one or four at test time.

\end{itemize}
During training, we construct each minibatch by randomly sampling 128 meshes from the relevant ShapeNet class uniformly with replacement.
For each selected mesh, we render a single image, using a pose sampled from $\text{Uniform}(-\pi,\,\pi)$.
Only these images are used to train the model, not the meshes themselves.
In experiments using multiple views, we instead sample 32 meshes and four poses per mesh, and correspondingly render four images.

\subsection{Generation}
\label{sec:gen-results}

\fig{samples} shows examples of meshes sampled from our model, using the same setting as \cite{gadelha173dv} (i.e. single-view training without pose supervision). That is the only prior work that learns a 3D generative model with just images as supervision.
We manually selected samples from our model that are stylistically similar to those from \cite{gadelha173dv} to allow side-by-side comparison.

We see that in all cases, generating meshes tends to give cleaner, more visually-pleasing samples than voxels (as used by \cite{gadelha173dv}).
For \textit{chair}, our model is able to capture the very narrow legs; for \textit{aeroplane}, it captures the diagonal edges of the wings; for \textit{car}, it captures the smoothly curved edges.
We have also successfully learnt a model for the concave class \textit{sofa}---which is impossible for \cite{gadelha173dv} as it does not consider shading.
Finally, note that our samples are diverse:
the model generates various different styles for each class.


\subsection{Reconstruction}
\label{sec:recon-results}

We now evaluate the performance of our model on 3D reconstruction from a single image.
We benchmark on a held-out test set, following the protocol of \cite{yan16nips}, where each object is presented at 24 different poses, and statistics aggregated across objects and poses.
We evaluate according to the following measures:
\begin{itemize}
\item
\textit{iou}: to measure the shape reconstruction error, we calculate the mean intersection-over-union between the predicted mesh and ground-truth; this follows recent works on reconstruction~\cite{yan16nips,tulsiani17cvpr}. To calculate this, we voxelise both meshes at a resolution of $32^3$
\item
\textit{err}: to measure the pose estimation error, we calculate the median error in degrees of predicted rotations
\item
\textit{acc}: again to evaluate pose estimation, we measure the fraction of instances whose predicted rotation is within $\pi/6$ of the ground-truth rotation.
\end{itemize}

\begin{table}[t]
  \small \centering
  \setlength{\tabcolsep}{2pt}
  \rowcolors{3}{gray!20}{white}
  \begin{tabular}{r|cccc|cccc|cccc|cccc}
    & \multicolumn{4}{c|}{\textbf{car}} & \multicolumn{4}{c|}{\textbf{chair}} & \multicolumn{4}{c|}{\textbf{aeroplane}} & \multicolumn{4}{c}{\textbf{sofa}} \\[-3pt]
    & \textit{iou} & \textit{err} & \textit{acc} & \textit{iou|$\theta$} &
    \textit{iou} & \textit{err} & \textit{acc} & \textit{iou|$\theta$} &
    \textit{iou} & \textit{err} & \textit{acc} & \textit{iou|$\theta$} &
    \textit{iou} & \textit{err} & \textit{acc} & \textit{iou|$\theta$} \\
    \hline
    \textit{ortho-block} & 0.71 & 7.3 & \textbf{0.84} & 0.74 & 0.41 & 9.2 & \textbf{0.69} & 0.49 & 0.30 & 7.9 & 0.73 & 0.24 & \textbf{0.59} & \textbf{7.3} & \textbf{0.94} & \textbf{0.69} \\
    \textit{full-block} & 0.54 & 6.5 & 0.82 & 0.63 & \textbf{0.46} & \textbf{4.6} & \textbf{0.69} & \textbf{0.51} & \textbf{0.51} & \textbf{4.4} & \textbf{0.89} & \textbf{0.57} & 0.39 & 9.1 & 0.70 & 0.68 \\
    \textit{subdivision} & \textbf{0.77} & \textbf{4.7} & \textbf{0.84} & \textbf{0.81} & 0.39 & 7.9 & 0.65 & \textbf{0.51} & 0.49 & 6.7 & 0.64 & \textbf{0.57} & 0.39 & 14.7 & 0.52 & 0.59
  \end{tabular}
  \\[4pt]
  \caption{
Reconstruction performance for four classes, with three different mesh parameterisations (\sect{generative}).
For each class, the first three columns are in the default setting of no pose supervision and correspond to the metrics in \sect{recon-results}; iou|$\theta$ is the IOU when trained with pose supervision. Higher is better for \textit{iou} and \textit{acc}; lower is better for \textit{err}. Experimental setting: single-view training, colour lighting, shading loss.
}
  \label{tab:class-vs-paramn}
\end{table}

\begin{table}[t]
  \small
  \centering
  \setlength{\tabcolsep}{2.1pt}
  \rowcolors{3}{gray!20}{white}
  \begin{tabular}{r|cccc|cccc|cccc|cccc}
    & \multicolumn{4}{c|}{\textbf{car}} & \multicolumn{4}{c|}{\textbf{chair}} & \multicolumn{4}{c|}{\textbf{aeroplane}} & \multicolumn{4}{c}{\textbf{sofa}} \\[-3pt]
    & \textit{iou} & \textit{err} & \textit{acc} & \textit{iou|$\theta$} &
    \textit{iou} & \textit{err} & \textit{acc} & \textit{iou|$\theta$} &
    \textit{iou} & \textit{err} & \textit{acc} & \textit{iou|$\theta$} &
    \textit{iou} & \textit{err} & \textit{acc} & \textit{iou|$\theta$} \\
    \hline
    \textit{colour} & \textbf{0.77} & \textbf{4.7} & \textbf{0.84} & \textbf{0.81} & \textbf{0.46} & \textbf{4.6} & \textbf{0.69} & \textbf{0.51} & \textbf{0.51} & \textbf{4.4} & \textbf{0.89} & \textbf{0.57} & \textbf{0.59} & \textbf{7.3} & \textbf{0.94} & 0.69 \\
    \textit{white} & 0.58 & 13.8 & 0.82 & 0.81 & 0.25 & 33.6 & 0.49 & 0.42 & 0.42 & 7.7 & 0.85 & 0.54 & 0.51 & 56.1 & 0.49 & \textbf{0.71} \\
    \textit{cl.+sil.} & 0.46 & 65.2 & 0.29 & 0.64 & 0.28 & 51.7 & 0.35 & 0.48 & 0.20 & 17.8 & 0.57 & 0.47 & 0.27 & 89.8 & 0.15 & 0.57
  \end{tabular}
  \\[4pt]
  \caption{
Reconstruction performance with different lighting and loss. \textit{colour} indicates three coloured directional lights with shading loss; \textit{white} indicates a single white directional light plus white ambient, with shading loss; \textit{cl.+sil.} indicates coloured lighting with only the silhouette used in the loss. Our model can exploit the extra information gained by considering shading in the loss, and coloured directional lighting helps further. Experimental setting: single-view training, best mesh parameterisations from \tab{class-vs-paramn}.
}
  \label{tab:lighting}
\end{table}

\begin{table}[t]
  \small
  \centering
  \setlength{\tabcolsep}{6pt}
  \rowcolors{3}{gray!20}{white}
  \begin{tabular}{r|cccc|cccc}
    & \multicolumn{4}{c|}{\textbf{car}} & \multicolumn{4}{c}{\textbf{chair}} \\[-3pt]
    & \textit{iou} & \textit{err} & \textit{acc} & \textit{iou|$\theta$} &
    \textit{iou} & \textit{err} & \textit{acc} & \textit{iou|$\theta$} \\
    \hline
    \textit{single-view} & 0.77 & 4.7 & 0.84 & 0.81 & 0.46 & 4.6 & 0.69 & 0.51 \\
    \textit{4-view train, 4-view test} & \textbf{0.83} & \textbf{2.6} & \textbf{0.94} & \textbf{0.86} & \textbf{0.51} & 4.7 & 0.72 & \textbf{0.55} \\
    \textit{4-view train, 1-view test} & 0.81 & 5.1 & 0.93 & 0.83 & 0.46 & \textbf{2.5} & \textbf{0.78} & 0.50
  \end{tabular}
  \\[4pt]
  \caption{
Reconstruction performance with multiple views at train/test time. Our model is able to exploit the extra information gained through multiple views, and can benefit even when testing with a single view. Experimental setting: best mesh parameterisations from \tab{class-vs-paramn}, colour lighting, shading loss.
}
  \label{tab:multi-view}
\end{table}

\begin{table}[t]
  \small
  \renewcommand*{\arraystretch}{0.95}
  \centering
  \setlength{\tabcolsep}{6pt}
  \rowcolors{1}{white}{gray!20}
  \begin{tabular}{>{\itshape}c >{\itshape}c >{\itshape}c|cccc}
    & \textbf{lighting} & \textbf{loss} & \textbf{car} & \textbf{chair} & \textbf{aeroplane} & \textbf{sofa} \\
    \hline
    DRC~\cite{tulsiani17cvpr} & white & silhouette & 0.73 & 0.43 & 0.50 & - \\
    DRC~\cite{tulsiani17cvpr} & white & depth & 0.74 & 0.44 & 0.49 & - \\
    PTN~\cite{yan16nips} & white & silhouette & 0.71 & 0.50 & 0.56 & 0.62 \\
    PTN, our images & colour & silhouette & 0.66 & 0.22 & 0.42 & 0.46 \\
    \hline
    ours & white & silhouette & 0.71 & 0.25 & 0.53 & 0.68 \\
    ours & white & shading & 0.79 & 0.44 & 0.54 & \textbf{0.69} \\
    ours & colour & shading & \textbf{0.83} & \textbf{0.51} & \textbf{0.57} & \textbf{0.69} \\
    \hline
    PSG~\cite{fan17cvpr} & white & 3D & \textit{0.83} & \textit{0.54} & \textit{0.60} & \textit{0.71}
  \end{tabular}
  \\[4pt]
  \caption{
Reconstruction performance (iou|$\theta$) in a setting matching \cite{tulsiani17cvpr,yan16nips} (multi-view training; best parameterisations from \tab{class-vs-paramn}), but with mesh output instead of voxels. \textit{PTN, our images} is running the unmodified public code of \cite{yan16nips} with its normal silhouette loss, on our coloured images. The final row shows performance of a state-of-the-art method~\cite{fan17cvpr} with full 3D supervision---note that our colour results are comparable with this, in spite of using only unannotated 2D images as supervision
}
  \label{tab:competitors}
\end{table}

\paragraph{Object classes and mesh parameterisations.}
\tab{class-vs-paramn} shows the performance of our model on four different classes, comparing the three mesh parameterisations of \sect{generative}.
This focuses on our default setting of colour lighting, shading loss, single-view training without pose supervision (columns \textit{iou, err, acc}); we also give \textit{iou} when trained with pose supervision (column \textit{iou}|$\theta$).
We see that different parameterisations are better suited to different classes, in line with our expectations.
Cars have smoothly curved edges, and are well-approximated by a single simply-connected surface; hence, \textbf{subdivision} performs well.
Chairs vary in topology (e.g. the back may be solid or slatted) and sometimes have non-axis-aligned surfaces, so the flexible \textbf{full-block} parameterisation performs best.
Aeroplanes have one dominant topology and include non-axis-aligned surfaces; both \textbf{full-block} and \textbf{subdivision} perform well here.
Sofas often consist of axis-aligned blocks, so the \textbf{ortho-block} parameterisation is expressive enough to model them.
We hypothesise that it performs better than the other more flexible parameterisations as it is easier for training to find a good solution in a more restricted representation space. 
This is effectively a form of regularisation.
%
%
Overall, the best reconstruction performance is achieved for cars, which accords with \cite{tulsiani17cvpr,yan16nips,fan17cvpr}.

The low values of \textit{err} (and corresponding high values of \textit{acc}) indicate that the model has indeed learnt to disentangle pose from shape.
This is noteworthy given the model has seen only unannotated 2D images with arbitrary poses---disentanglement of these factors presumably arises because it is easier for the model to learn to reconstruct in a canonical frame, given that it is encouraged by our loss to predict diverse poses.
However, providing the ground-truth poses as input improves reconstruction performance further in almost all cases (column \textit{iou}|$\theta$ vs. \textit{iou}).

\paragraph{Benefit of lighting.}
\tab{lighting} shows how reconstruction performance varies with the different choices of lighting, \textbf{colour} and \textbf{white}, using \textbf{shading} loss.
Coloured directional lighting provides more information during training than white lighting, and the results are correspondingly better.
We also show performance with \textbf{silhouette} loss for coloured light.
This performs significantly worse than with shading in the loss, in spite of the input images being identical.
Thus, back-propagating information from shading through the renderer does indeed help with learning---it is not merely that colour images contain more information for the encoder network.
%
As in the previous experiment, we see that pose supervision helps the model (column \textit{iou}|$\theta$ vs. \textit{iou}).
In particular, only with pose supervision are silhouettes informative enough for the model to learn a canonical frame of reference reliably, as evidenced by the high median rotation errors without (column \textit{err}).

\paragraph{Multi-view training/testing.}
\tab{multi-view} shows results when we provide 4 views of each object instance to the model.
Using 4 views at both training and testing time improves results in all cases---the model has learnt to exploit the additional information about each instance.
There is also a smaller performance improvement when we train with 4 views, but test with only one---although the network has not been optimised for the single-view task during training.

\paragraph{Comparison to previous works.}
\tab{competitors} compares our results with previous works.
Here, we conduct experiments in a setting matching \cite{tulsiani17cvpr,yan16nips}: multiple views at training time, with ground-truth pose supervision.
This shows that our results using meshes are roughly comparable with these previous works using voxels, even when only silhouette supervision is used (our results are worse on `chair', but better on `sofa').
Furthermore, when we add shading information to the loss (which these previous works cannot), our results show a significant improvement; coloured lighting helps even further.
We also show results for \cite{yan16nips} using our coloured lighting images as input, but their silhouette loss.
This performs worse than our method on the same images, again showing that shading in the loss is useful---our colour images are not simply more informative to the encoder network than those of \cite{yan16nips}.
Interestingly, when trained with shading or colour, our method outperforms \cite{tulsiani17cvpr} even when the latter is trained with depth information.
When trained with colour, our results are even close to \cite{fan17cvpr}, which is a state-of-the-art method trained with full 3D supervision.

\section{Conclusion}

We have presented a framework for generation and reconstruction of 3D meshes.
Our approach is flexible and supports many different supervision settings, including weaker supervision than any prior works (i.e. a single view per training instance, and without pose annotations).
Unlike prior works, we can exploit shading cues due to directional lighting; we have shown that this improves performance over silhouettes.
Moreover, performance is higher than that of a method with depth supervision~\cite{tulsiani17cvpr}, and even close to the state-of-the-art results using full 3D supervision~\cite{fan17cvpr}.
Finally, ours is the first method that can learn a generative model of 3D meshes, trained with only 2D images.
We have shown that use of meshes leads to more visually-pleasing results than prior voxel-based works~\cite{gadelha173dv}.

\bibliography{../../bibtex/shortstrings,../../bibtex/calvin,../../bibtex/vggroup}

\end{document}